\documentclass{article}

\usepackage{microtype}
\usepackage{graphicx}
\usepackage{subfigure}
\usepackage{booktabs} % for professional tables

\usepackage[linesnumbered,ruled]{algorithm2e}
\usepackage{parskip}
\usepackage[accepted]{icml2023}

\usepackage[colorlinks,citecolor=blue,urlcolor=red,bookmarks=false,hypertexnames=true, linkcolor=green]{hyperref}

\usepackage{amsmath}
\usepackage{amssymb}
\usepackage{mathtools}
\usepackage{amsthm}

\usepackage[capitalize,noabbrev]{cleveref}

\theoremstyle{plain}

\theoremstyle{definition}

\theoremstyle{remark}

\usepackage[textsize=tiny]{todonotes}

\icmltitlerunning{}

\begin{document}

\twocolumn[
\icmltitle{FIMBA: Evaluating the Robustness of AI in Genomics via Feature Importance Adversarial Attacks}



\icmlsetsymbol{equal}{*}

\begin{icmlauthorlist}
\icmlauthor{Heorhii Skovorodnikov}{main}
\icmlauthor{Hoda AlKhzaimi}{main}
% \icmlauthor{Firstname3 Lastname3}{comp}
\end{icmlauthorlist}

\icmlaffiliation{main}{New York University}

\icmlcorrespondingauthor{Heorhii Skovorodnikov}{hs3470@nyu.edu}

\icmlkeywords{Machine Learning, ICML}

\vskip 0.3in
]



\printAffiliationsAndNotice{}  % leave blank if no need to mention equal contribution

\begin{abstract}
With the steady rise of the use of AI in bio-technical applications and the widespread adoption of genomics sequencing, an increasing amount of AI-based algorithms and tools is entering the research and production stage affecting critical decision-making streams like drug discovery and clinical outcomes. This paper demonstrates the vulnerability of AI models often utilized downstream tasks on recognized public genomics datasets.
We undermine model robustness by deploying an attack that focuses on input transformation while mimicking the real data and confusing the model decision-making, ultimately yielding a pronounced deterioration in model performance. Further, we enhance our approach by generating poisoned data using a variational autoencoder-based model. Our empirical findings unequivocally demonstrate a decline in model performance, underscored by diminished accuracy and an upswing in false positives and false negatives. Furthermore, we analyze the resulting adversarial samples via spectral analysis yielding conclusions for countermeasures against such attacks.
\end{abstract}

\section{Introduction}\label{sec1}
Adversarial machine learning (AML) has become a widely recognized field in recent years due to its potential to test, validate and manipulate AI models. With the growing use of AI in genomics and clinical applications, the need for robust and secure models has become increasingly important. Genomic data plays a crucial role in diagnosing diseases in patients. However, the use of AI models in genomics is not without its challenges, particularly in the case of their security. The dangers of tampering or hacking machine learning pipelines in genomics and clinical applications can introduce risks that cannot be overstated, as it can potentially cause significant harm to patients and the public at large \cite{finlayson2019adversarial}.
For example, in recent years, genomics sequencing algorithms have become more accessible and affordable, making them widely used in clinical settings
\cite{yang2020review}. The importance of genomics sequencing algorithms in disease diagnosis lies in their ability to identify genetic mutations that may cause or contribute to the development of certain diseases. This information can be used to develop personalized treatment plans and potentially prevent the onset of certain diseases in at-risk individuals. Overall, genomics sequencing algorithms have become an essential tool in modern medicine, and their usage is expected to continue to increase in the coming years as the field of genomic medicine continues to advance \cite{jacob2013genomics}.
Among them machine learning plays an integral part in various ways of data processing and downstream analysis, significantly increasing the speed of data acquisition and enabling academic and industry innovation \cite{whalen2022navigating}. As a subset of machine learning, deep learning also plays a vital role in utilizing genomics data for multiple vital tasks, like genome analysis, disease detection, classification, and anomaly detection \cite{zou2019primer, xu2019translating}.
 However, while the pace of research for applications increased, studies on security risks and model robustness remain largely unexplored. To our knowledge, our study is one of the first and early studies to provide a robustness analysis for AI models in genomics. We hope to fill a gap in the field of AI robustness when it comes to genomics algorithms and provide foundational research that can be used as a base study for further works within the community.
\subsection{Related work}

To understand the demand for a robustness study, we performed a literature review to identify popular architectures as well as those that are likely to be used in the near future. We identified a wide usage of Random Forest (RF) in literature \cite{chen2012random,nguyen2015genome} for different kinds of genomic tasks. Additionally, the tree-based XGBoost method is a popular choice for classification and prediction tasks using genomic data \cite{li2017comprehensive,li2019gene,zhong2018xgbfemf}. We investigated the use of deep learning (DL) as well as it is a prevalent choice in the industry and DL models are often either used as standalone architectures or combined in an ensemble pipeline together with other machine learning (ML) models, depending on the downstream task. Specifically for genomics data, we see the rise of popularity of convolutional-based models, e.g. CNN \cite{liu2020application,kuang2022accurate}, mainly due to the fact that CNNs are widely considered to be a good feature extractor and can work on high dimensional data. However, while the usage of deep learning and adoption of advanced architectures is increasing, the research is still in its nascent stages. Hence, in our study, we additionally focus on powerful models that are often used in computer vision and NLP that have recently found their way into genomics, namely residual nets \cite{ye2021genomic,nair2019integrating} and transformers \cite{clauwaert2020novel,zaheer2020big,le2022deep,ji2021dnabert}. Given rapid academic adoption, we conclude that the industry and scientific impact is not far behind, this fact coupled with a plethora of serious risks and consequences of erroneous AI in genomics further motivates the necessity of our study.

The pace of robustness research lags significantly, as we found a smaller number of studies that looked at different aspects of the proposed AI detectors and classifiers. \cite{kuo2022evolving} note that the biggest concern in the research community is the security of anonymity of genomic data, this is supplemented by the fact that a lot of such data can be publicly accessed and popular services like 23andMe are routinely used. Privacy-preserving ML techniques such as differential privacy are being attacked to extract the identifying information of a subject's genome \cite{chen2020differential,humbert2015anonymizing}, additionally \cite{huang2015genoguard} focus on preventing access to data from adversarial access. However, this demonstrates the lack of research into adversarial attacks on detectors themselves with the goal to lower their accuracy and misclassify a prediction. We specifically focus on a classification task, as identifying populations or diseases is a highly lucrative avenue for both attackers and industry players. Here understanding of model behavior is crucial, \cite{koo2019robust} showed that DL models suffer from low interoperability, and thus have a lack of transparency. 
Recently, many methods have been proposed as attacks in the literature targeting models operating on images, text, audio, and video \cite{li2022review}. Attacks vary from black-box \cite{chen2017zoo,tavallali2022adversarial} and white-box \cite{carlini2017adversarial,goodfellow2014explaining} versions. However, adversarial attacks on genomic data remain largely unexplored. 
\cite{jain2021generator} focuses on a set of methods to perform clustering and find regions volatile to mutation in genes. While this paper presents an interesting method of exploring genomic sequences via state machines. It does not directly apply to an adversarial space. The framework focuses on the task of improving the understanding of genomic sequence and the detection of natural mutations, these sequences come in the form of text. 
While our gene expression data is given in numerical values via gene counts, models trained on its continuous representation are more similar to the computer vision task of object classification. Further, in comparison, our paper directly focuses on an attack against a suite of commonly used models. Our method doesn't aim to resemble a natural mutation directly, rather we attempt to mask it as a normal sample while targeting features vulnerable to the model itself as each model has a different understanding of the data.
For the detection of changes, we apply a more commonly used and robust method relying on a fast Fourier transform. Additionally, we test our assumptions on two large and widely used datasets with critical applications.
\cite{cheng2021rumor} attempts to detect textual anomalies in tweets via a GAN-based method yet performs a small experiment on the detection of gene mutations. We find the experiments on detection are largely limited and are not geared towards defense against adversarial attacks, instead, they are focused on improving gene mutation classification tasks. Our method in contrast is adversarial and focuses on continuous representations (which are notoriously harder to defend against) and their undetectability.

The closest to our work is a very recent study by \cite{mas2022adversarial} that demonstrated the ability of a gradient-based white-box attack on a deep learning detector in genomics. In our paper we introduce Feature Importance Black-Box Attack(FIMBA), in contrast to previous methods, we employ more advanced architectures, utilize a straightforward and simple black-box approach, and analyze the results of the attack for detectability and countermeasures.

To the best of our knowledge, this work is the first to demonstrate black box attacks on gene expression data, we are also the first to attack many popular models that are often utilized in literature and industry.
The main contributions of this paper can be  highlighted by the following: 
\begin{itemize}
    \item We introduce a novel framework of attack and analysis for popular machine learning models in genomics, that utilizes a distinct approach to affect the final prediction of the model. All our methods can be considered black-box and model-agnostic as they do not require access to the model gradient.
    \item We study the success rate of the attack at varying intensities and conclude the best possible attack settings. On the basis of our results, we discuss countermeasures aimed at strengthening popular architectures. 
    \item We supplement our previous attack methods with a generative attack model capable of rapidly building poisoned synthetic datasets ready to be inserted into AI-powered genomics pipelines.
\end{itemize}

\section{Methodology: Attack Pipeline}
\subsection{Selected Datasets}\label{sec:data}
To make our analysis robust and as close to real-world situations as possible we decided to select well-known datasets that target critical applications. A search was performed online via Google Scholar to find commonly used sources of data. Platforms included Terra, NCBI-GEO, Human Cell Atlas, and UCSC Xena. A common factor in genetic datasets is a high dimensionality of data while possessing a low individual sample count, as for model training a bigger dataset size is preferred, we performed the search to filter publicly available datasets. In this work, we focus on two critical applications of genomic data: detection of cancer and COVID-19. The task is classification with the goal of identifying whether the sample is malignant or not for the cancer dataset, and whether a patient has COVID-19. The processed dataset card is shown in table \ref{dataset-card}.

\begin{table}[h]
  \centering
  \begin{tabular}{|l|l|l|l|}
    \hline
    Name     & Negative   & Positive  & Total \\
\hline
    \textbf{TCGA}      & 8167  & 10506  & 18673    \\
    \textbf{COVID} & 15767 & 34233 & 50000 \\
    \hline
  \end{tabular}
    \caption{Dataset card}
  \label{dataset-card}
\end{table}

Additionally, for this study, we decided to focus on a common sequencing type: single-cell RNA sequencing. A general idea of operating on such data is that a model can extract meaningful representations of a class by looking at a numeric count of a specific gene in a sample. 
For our sample of viral diseases, we selected Columbia University/NYP COVID-19 Lung Atlas from NCBI GEO databank with accession ID GSE171524. The data was sequenced using Illumina NovaSeq 6000 platform and primarily consists of 116,314 cells taken from 19 COVID-19 decedents and seven control patients with short postmortem interval (PMI) autopsies.\cite{melms2021molecular}.
For cancer prediction we used one of the largest and most comprehensive datasets available to researchers online: The Cancer Genome Atlas (TCGA) \cite{tomczak2015review}; a landmark cancer genomics program, that molecularly characterized over 20,000 primary cancer and matched normal samples spanning 33 cancer types. 
A limitation of TCGA data was the lack of negative samples, therefore we saw fit to expand these datasets with two others Therapeutically Applicable Research To Generate Effective Treatments (TARGET) and Genotype-Tissue Expression (GTEx) \cite{gtex2015genotype}. The TARGET dataset collects pediatric data on different cancers and is considered to complement TCGA in providing a more comprehensive genetic profile of different cancers. GTEx dataset contains samples of normal, non-cancerous cells, this allows for a better comparison between different gene expression profiles and helps to remedy shortcomings in non-cancerous samples from the TCGA dataset as the negative samples there were taken from areas near the cancer cells, potentially causing some overlap in their expression. We refer to these datasets as "TCGA" and "COVID-19".
Datasets were normalized in the range between $R=(-1,1)$ using min-max scaling and split in 67/33 proportion for training and testing respectively, with random seed $= 42$. The two datasets were also selected for their varying degree of complexity, as TCGA and COVID datasets given PCA applied to them, showed different levels of complexity with the latter being more complex and therefore more challenging to classify. Complex data is more likely to be encountered in industry settings and therefore it is important to demonstrate the attack influence on varying dataset types.
The detailed procedure for processing datasets as well as clustering analysis are located in our data appendix.
\subsection{Models and Compute Infrastructure}

Based on the existing literature and our projection we selected five architectures to act as our classifiers: Random Forest, XGBoost, CNN, ResNet, and Vision Transformer.
We trained them using the following model configurations:
\begin{itemize}
    \item \textbf{Random Forest}: estimators = 100, Gini impurity criterion, nodes are expanded until maximum node purity is reached or until there is a single sample per node.
    
    \item \textbf{XGBoost}: estimators = 100, booster = 'gbtree',  
learning rate = 0.3,   maximum depth = 6.
    
    \item \textbf{CNN}: Conv2D layers = 3, fully connected layers = 2, loss = BCE , learning rate =0.001, optimizer = Adam, epochs = 30.
    
    \item \textbf{ResNet18}: Modified ResNet18 with improvements \cite{molchanov2016pruning}, loss = BCE, learning rate = 0.001 , optimizer = Adam, epochs = 15.
    
    \item \textbf{ViT}: Conv2D layers with batch normalization = 3, fully connected layers = 2, transformer block = 1, loss = BCE , learning rate = 0.001, optimizer = SGD, epochs = 25.
\end{itemize}
For RF and XGBoost, the training was done on the 20 AMD EPYC 7742 CPUs, parallelized for speed, inference was performed on a single CPU of the same type. For deep learning models, the training and inference were performed on a single NVIDIA A100 GPU. Models were developed using PyTorch. The attacks were written in Python and deployed on a Linux server.
We used random seed  $r=42$ for the training and evaluation of classifiers. Epochs indicate training epochs the model was trained for. For Conv2D layers, we used the standard practice of forming an image out of the feature vector by getting the vertical and horizontal dimensions as even as possible. Here by a single attack, we mean running each method and model using $n$ amount of features as a target for modification.

\subsection{Proposed Attack Approach}
The center theory behind the attack relies on two main components: model decision space, and feature importance. It must be noted that while we perform our attack on the whole test dataset to quantitatively measure its success, in real-life scenarios, changing just one patient's record can lead to devastating consequences for both the patient and the provider.  A core assumption is that for a select set of features for each feature, the model makes a decision on a boundary. This separation can be expressed as a high dimensional hyperplane $L$ of the decision space. Mapping the distance $D$ between the two closest samples $X_0$ and $X_1$ of the opposite class dictates the space in which for a specific value 
of a feature a different model outcome can be made based on the relative position on that line. Given the model weights a probabilistic space occurs, between potential two points where $f_0$ and $f_1$, represent the feature values. Mapping that space shows the potential model decision given different values of a feature $f$.  We utilize this phenomenon to extract a new value for an attacked feature. We wanted to avoid using a model gradient, as often, given, the privacy risks, the access to model parameters can be safeguarded. We note that false positive and false negative samples (FP/FN) are our main target as with original adversarial attacks the goal is to keep the modified sample as close to the original as possible so as to avoid detection  \cite{goodfellow2014explaining}. While in the case of images, the visual inspection can show a large scale of distortion, it is harder to observe genomics samples in the same way, nevertheless as the feature vector, in this case, can be interpreted as a signal, tools to detect image attacks can be well applied here as well. Therefore, the motivation for lesser detectability still stands. Ideally, our attack sample would mimic FP/FN to masquerade as a natural model error.
Additionally, we wanted to investigate the ways to make our attack less detectable, therefore we assume the benefit of transforming fewer features. This is akin to modifying fewer pixels during the attack on the image classifiers. To this end, we experimented with SHAP values in order to select the most susceptible features \cite{lundberg2017unified}.  In practice, a different feature importance extraction method can be used e.g. LIME. We chose SHAP for its popularity and relative ease of implementation. The algorithm behind the attack is described in algorithm table \ref{mainatk-algo}. As we don't access model parameters we don't consider this attack white-box using traditional definition. However, we do assume access to the dataset, this assumption, to note, is less strict as we find that not only the attackers can use widely available public datasets but also use methods to generate synthetic data as we also demonstrate below, we hence conclude that in practice full data access might not always be needed for a successful attack.
We begin by simulating potential detectors via pretraining them as our baseline. We train our classifiers and evaluate their baseline performance on two datasets. During the attack, we vary the severity of the attack by changing the number of features affected while collecting accuracy scores and the number of false negatives and false positives (FN/FP).

\begin{algorithm}[tb]

\caption{Main Attack Algorithm}
\SetAlgoLined
\label{mainatk-algo}
\SetKwInOut{Input}{Input}
\SetKwInOut{Output}{Output}

\Input{ $\{V_f\}$ - feature vectors to modify}
\Output{$\hat{D}$ - modified dataset}

$V_f:\{V_{0(FP)},...\}\lor\{V_{0(FN)},...\}$ // Select viable feature vectors

$V*: \{v_0,...,v_n\}$ // Select vector to attack

$i \leftarrow D_s $: Dataset Size 

\While{$c_{i-1} \neq D_s$}{
$\{V_{near}\} \leftarrow d_{Minkowski}(V*, \{V_f\})$ // Compute similarity 

$\big\downarrow{\{V_{near}\}}$ //  Sort descending

$v_i* \leftarrow I(v_i, V_{fp} \lor V_{fn}, n)$ : $(Fscore | N_f)$

$\hat{D} \leftarrow v_i*$

$i \leftarrow i + 1$
}

return

\end{algorithm}

We take a vector or a set of vectors to be modified $V*$ and a corresponding set of target vectors. To draw a parallel with the computer vision domain and pixel attacks, it would be similar to choosing an image that is more susceptible to model confidence reduction and therefore was misclassified, e.g. our FP/FN samples. We then compute the measure of similarity between $V*$ and $V_f$, using the Minkowski distance given in equation \ref{dist-eq}, after sorting the results to give us the closest member of $V_f$ to our $v | V*$, this is essentially equivalent to nearest-neighbor search in high dimensional space. To optimize this step we use KDtrees \cite{ram2019revisiting} given their speed in higher dimensions, however, any search algorithm with similarity measure can be suitable. 
The Minkowski distance between two vectors $\mathbf{v_1}$ and $\mathbf{v_2}$ of length $n$ is defined as:
\begin{equation}\label{dist-eq}
    d_{Minkowski}(\mathbf{v_1}, \mathbf{v_2}) = \left(\sum_{i=1}^n |v_{1,i} - v_{2,i}|^p\right)^{\frac{1}{p}}
\end{equation}
Where $p$ is a positive integer parameter that controls the degree of the distance metric, 
$p = 1$ is equivalent to the Manhattan distance, $p = 2$, is equivalent to the Euclidean distance. At the same time, we want to perform the jump-along feature in a certain direction. As we show below, we found that no single feature is highly damaging to the model, instead it is a combination of these features that produce a great effect, therefore this motivates the feature selection step.
To select a feature to modify, we want to pick the most impactful ones as fewer modification results in a more efficient attack and will be harder to detect. 
To perform this selection we utilize the SHAP-based method, which is model agnostic and relies on repeated feature permutation to extract the importance scores described in equation \ref{shap-eq}. 
Additionally, it doesn't require the knowledge of model internals and therefore is valid for a black-box approach.
SHAP values are used to explain the prediction of a machine learning model by attributing the contribution of each feature to the prediction. 

\begin{equation}\label{shap-eq}
    \phi_{i,j} = \sum_{S \subseteq N \setminus {j}} \frac{|S|! (m - |S| - 1)!}{m!} [f(x_i | x_{i,S}) - f(x_i | x_{i,N \setminus {j}})]
\end{equation}

Where $N$ is the set of all features, $m$ is the number of features, $f(x_i | x_{i,S})$ is the expected value of the model's prediction for sample $x_i$ given the features in set $S$, 
and $f(x_i | x_{i,N \setminus {j}})$ is the expected value of the model's prediction for sample $x_i$ given all features except feature $j$.
The SHAP values $\phi_{i,j}$ represent the contribution of feature $j$ to the prediction for sample $x_i$. 
We found that a set of most important features to the model computed over the whole dataset and over the sample are practically similar in their sorted order. Hence SHAP values for the attack on the whole dataset can be computed over a few samples, thus reducing number of queries to the model. Therefore for our attack we tried both computing the values for each sample separately and precomputing the values globaly on a small set of samples, the success rate for both variants was found to be without significant statistical difference.  For efficiency we recommend sampling a small set of feature vectors with different final outputs and computing the SHAP values, we found sample size of 10 sufficient.
Local computation is better for a targeted attack and as deriving SHAP values can be computationally expensive to do for whole datasets, a few sample computation is beneficial in a way of efficiency.

After selecting which features to modify we interpolate between our target vector and our original vector to produce a third attack vector using equation \ref{intp-eq}. Specifically, we do so on the features we picked using equation \ref{shap-eq}.
\begin{equation}\label{intp-eq}
    {I}(a, b, n) = \left[ a + i \cdot \frac{b-a}{n-1} \right]_{i=0}^{n-1}
\end{equation}
Additionally, parameter $n$ allows us to regulate the sampling or 'granularity' of interpolation. For our experiment, if we consider interpolation direction $a \rightarrow b$, where out of $n$ resulting vectors, $a$ is the first one in the ordered set and $b$ last as our interpolation endpoint, we take vector at position $n-1$ for maximum similarity. For our experiments, we set $n=10$. Therefore, this approach also doubles as a generation method and allows us to generate vectors of varying severity, hence, we can further regulate the strength of the attack. The resulting complexity of this attack is $O(nkplog(m))$, where $n$ is the number of samples, $k$ is the number of features you are modifying for each sample, $p$ is a precision value for interpolation, and $log(m)$ is a cost of search of K-D tree. We experimented with using FP/FN and true positives/true negatives (TP/TN) as our interpolation endpoints both. Hence, $a$ and $b$ represent transformation pairs, arranged with a goal to illicit model confidence reduction and label change.

\begin{table*}[htbp]
\centering
\begin{tabular}{|c|c|c|c|c|c|c|}
\hline
\multicolumn{4}{|c|}{\textbf{TCGA Data}} & \multicolumn{3}{c|}{\textbf{COVID Data}} \\
\hline
\textbf{Model} & \textbf{SSIM(Dataset)} $\uparrow$ & \textbf{SSIM(FP)} $\uparrow$ & \textbf{SSIM(FN)}  $\uparrow$& \textbf{SSIM(Dataset)} $\uparrow$& \textbf{SSIM(FP)} $\uparrow$& \textbf{SSIM(FN)} $\uparrow$\\
\hline
RF-m & 0.903 & 0.836 & 0.479 & 0.751 & 0.983 & 0.095  \\
XGBoost-m & 0.907 & 0.866 & 0.349 & 0.971 & 0.929 & 0.704 \\
CNN-m & 0.893 & 0.810 & 0.708 & 0.992 & 0.986 & 0.664  \\
ResNet-m & 0.905 & 0.807 & 0.661 & 0.844 & 0.995 & 0.657 \\
ViT-m & 0.904 & 0.929 & 0.709 & 0.978 & 0.989 & 0.887  \\
\hline
RF-r & 0.004 & 0.006 & 0.003 & 0.004 & 0.156 & 0.015 \\
XGBoost-r & 0.004 & 0.005 & 0.003 & 0.003 & 0.009 & 0.006\\
CNN-r & 0.004 & 0.005 & 0.003 & 0.0407 & 0.080 & 0.127 \\
ResNet-r & 0.004 & 0.010 & 0.021 & 0.010 & 0.042 & 0.042 \\
ViT-r & 0.005 & 0.009 & 0.002 & 0.016 & 0.689 & 0.123 \\
\hline
\end{tabular}
\caption{Spectral analysis of the attacks demonstrating undetectability, "-r" suffix indicates brute force random attack, "-m" main approach. $1.0$ is the maximum possible and desired value, indicating a perfect sample structure match. SSIM is computed between original and attacked versions of data, FP/FN indicates subsets of a dataset that consist only of those samples.}
\label{methodcomparison}
\end{table*}

\subsubsection{Ablation: Brute Force Approach}
Two key components of our attack are the selection of features to modify which in its original form is carefully done to mimic natural model errors (FP/FN), and the choice of the value by which a selected feature would change. To study the tradeoffs and effectiveness of our original method, we also experimented by randomizing these two parts for our attack.  Here we completely drop the SHAP feature component and instead sample feature indexes from the discretized uniform distribution, by rounding down the value within the range of $(0, N_{features})$. Our goal was to see the brute force effectiveness against other approaches on both datasets. The feature modification values are sampled within the range $(-1,1)$, the same as the normalization range of the dataset, avoiding the interpolation step. The resulting complexity of this attack is $O(nk)$, where $n$ is the number of samples, $k$ is the number of features being modified for each sample.

\subsubsection{Generation of Poisoned Data}
To study the possibility of insertion of synthetic poisoned data into existing pipelines as well as to address the potential limitation of access to data for the attack we investigated a way to use generative models in order to create a fake FP/FN sample that was inspired by our main method. For this task, we developed a variational autoencoder (VAE),  encoder, and decoded architectures that are specific to each dataset in that the input and output dimensions are regulated by original feature vector dimensions. All models have 6 layers for encoder and decoder each, consisting of the following structure of linear layers $In_d \rightarrow 2100 \rightarrow 1600 \rightarrow 1200 \rightarrow800 \rightarrow 512 \rightarrow l_d$  ($In_d$ is input dimension; $l_d$ latent dimension, decoder structure is a reverse of this), and a sampler in the middle. 
Next, we trained our VAE for 200 epochs using reconstruction loss. Our model was trained on two Nvidia V100 GPUs. The training procedure is the same for both datasets, resulting in two separate, tuned generative models. Since VAE is a probabilistic model the loss we employed for training is a combination of mean squared error (MSE) as our reconstruction loss and Kulback-Leibler divergence (KLD) as the regularization term. The intuition here is MSE+KLD addresses both  encoding-decoding scheme improvement and regularization of latent space, it is expressed as:
$$
Loss_{VAE} = ||x-x^|||^2+ KL(N(\mu_x, \sigma_x)|N(0,1))
$$
where N is a Gaussian distribution, $\mu$, and $\sigma$ represent feature mean and deviation respectively. 
For our attack we use a similar method to our original approach described above, we encode original and target vectors and perform interpolation in their encoded form, and sample vectors for decoding next to the endpoint which is our target sample.  This greedy method addresses the potential lack of access to data and as we demonstrate in the results section, we are able to use VAE to create samples that closely match the original, the noise induced by latent space can be considered beneficial here as it provides a degree of diversity to the synthetic sample.

\section{Results}

\begin{figure*}
    \centering
    \includegraphics[scale=0.56]{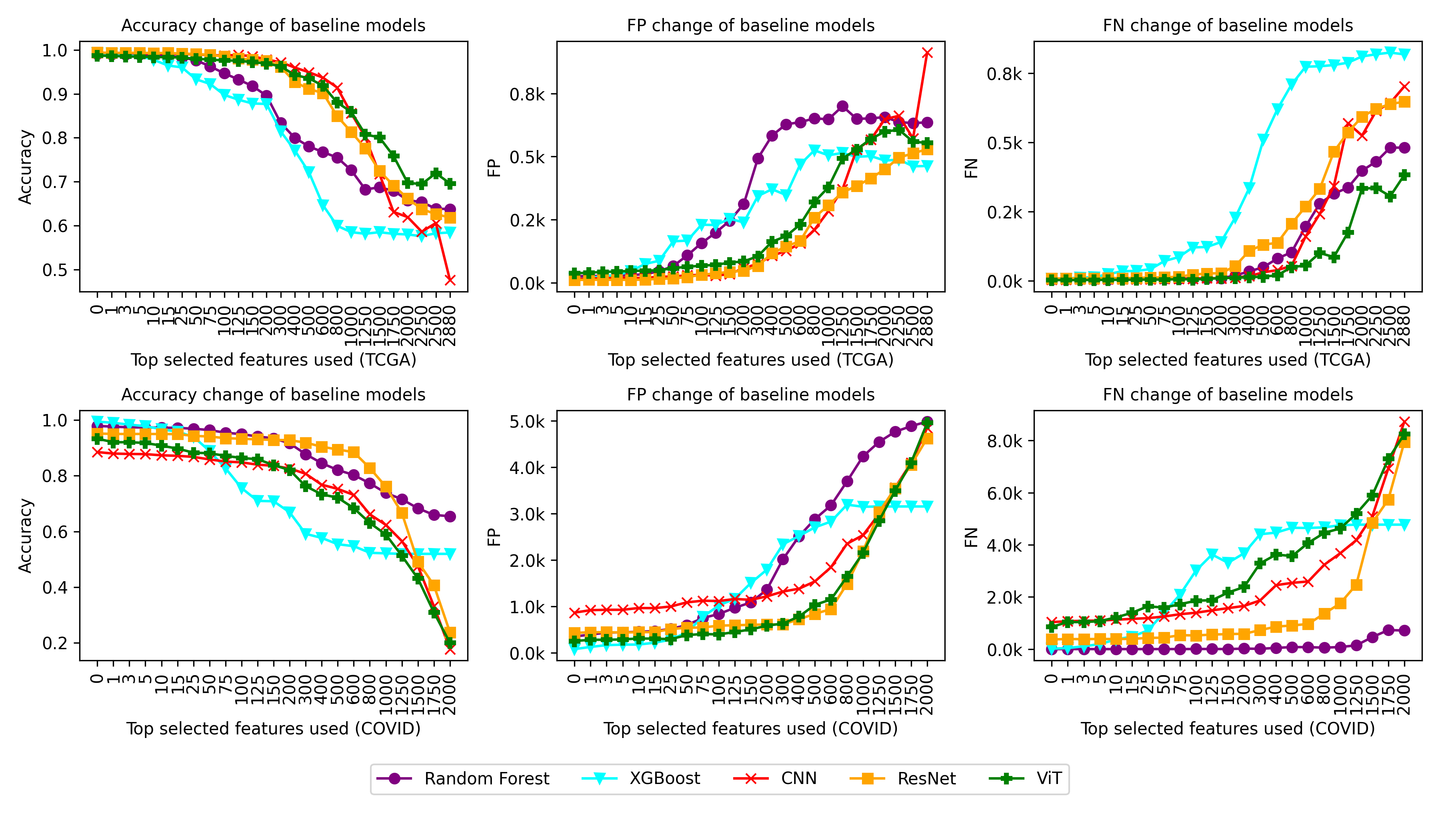}
    \caption{Results of our main attack. $k=1000$ for the number of samples; x-axis represents different attack strength settings via the number of features modified}
    \label{fig:atk1}
\end{figure*}

\begin{figure*}
    \centering
    \includegraphics[scale=0.56]{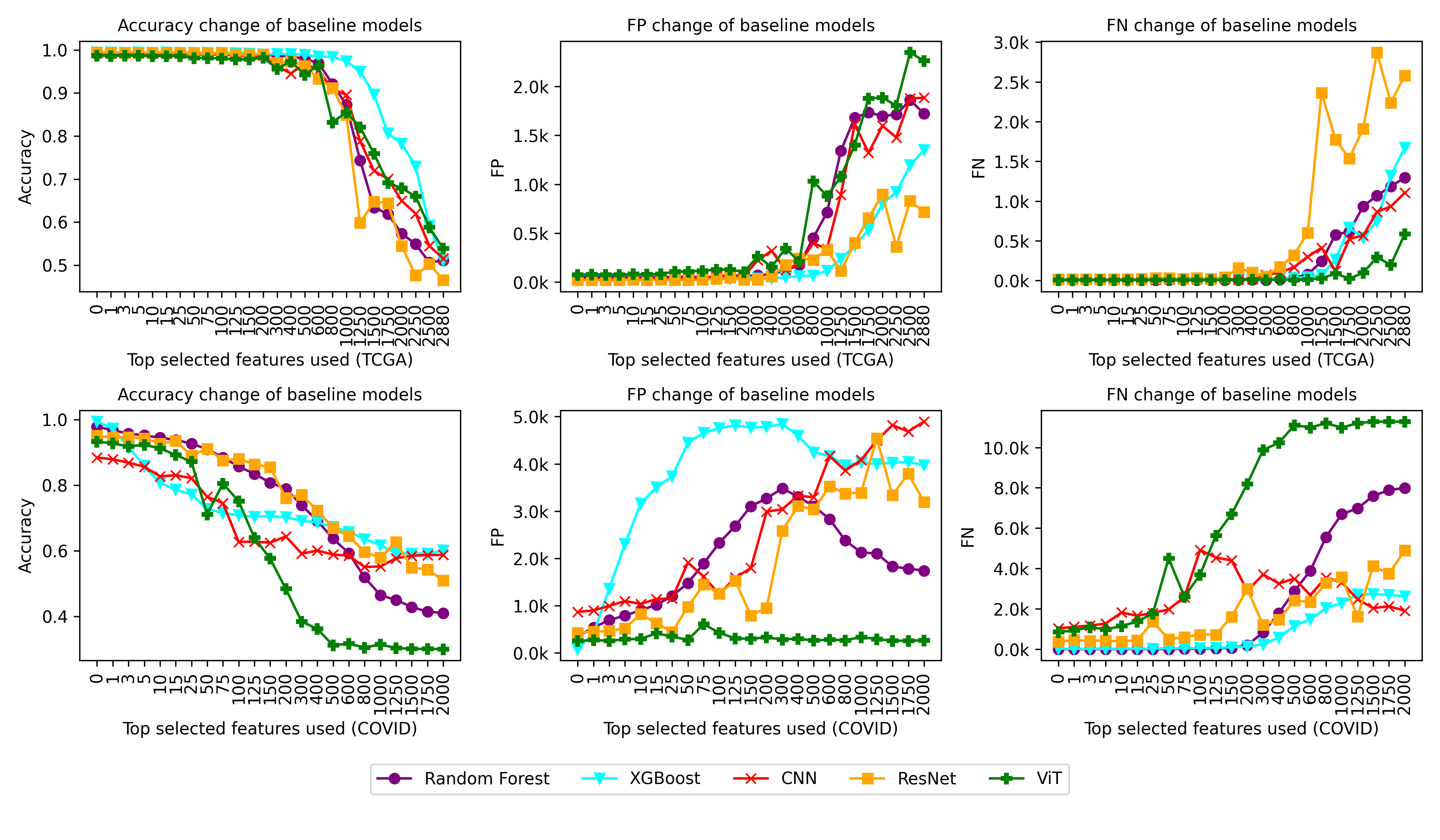}
    \caption{Results of our brute force attack. k = 1000 for the number of samples; x-axis represents different attack strength settings via the number of features modified}
    \label{fig:atkrand}
\end{figure*}
The effectiveness of our attacks can be seen in figures \ref{fig:atk1}\&\ref{fig:atkrand}. We note clear effectiveness in both the decrease in performance and increase in FP/FN samples, our main method elicits effect early while both methods show exponential destructivity after 400 features transformed mark. This value is different between COVID-19 and TCGA datasets for both attacks, we must note that the COVID-19 dataset possessed a more complex distribution between positive and negative classes, this in turn results in a complex separation boundary but also seemingly makes models trained on complex datasets more vulnerable for modification attacks even on smaller scales. Another notable discrepancy between the two methods is a controlled and stable generation of FP/FN. For the brute force approach on a more complex COVID-19 dataset, we observe peaks and falls for both FP and FN samples, in general, if the FP amount goes up FN goes down and vice versa. Tree-based models also seem to have better resistance to selective feature modification, but they also fail strongly on brute-force attacks. 
Even though the number of FP samples occurring was high for both XGBoost and RF models, the accuracy decrease plateaued at 500-600 features for the TCGA dataset and 800-1250 on the COVID-19 dataset, compared to deep learning models that for both datasets demonstrated a stable drop in accuracy.

Overall both attack formats demonstrate great effectiveness. For larger subsets of features under attack it would make more practical sense to go with brute force for higher efficiency, but as we show below this approach has a caveat. The other central component of our analysis is the detectability of the attack. As we demonstrate in our section on countermeasures below, while both attacks are effective, we note significant differences in the transformed sample structure. Recalling that the goal of AML is to optimize for both effectiveness via destructivity and undetectability, there can occur a tradeoff between these two objectives. We measure how undetectable our attacks are by determining how close the transformed sample is to the natural, original model error. As visual inspection of genomics data is impractical, we apply other statistical forms of analysis to determine a difference in resulting sample patterns.

\subsection{Comparison to Other Methods}

To provide a more complete overview of the benefits and tradeoffs of our method we compared it against a selection of popular adversarial attacks in the field. Since our approach focuses on a new type of black-box attack we wanted to compare mainly against other black-box approaches, specifically we focus on two attacks: a pixel attack \cite{su2019one} and Zeroth-Order-Optimization (ZOO) \cite{chen2017zoo}.
On deep learning models, we run extra comparisons with white-box attacks, specifically, we ran Fast Gradient Sign Method (FGSM) \cite{goodfellow2014explaining} and Auto Projected Gradient Descent (Auto-PGD) \cite{croce2020reliable} on deep learning models.

For the implementation, we attempt to match the original papers in terms of the hyperparameters and algorithm of the attack. However, as one of the goals of our attack is efficiency, we modified some attacks to produce better and faster results. Attacks were developed with the help of IBM ART \cite{art2018}.
FGSM run utilized step size $\epsilon_{step} = 0.3$ and batch size of 32, with maximum perturbation value $\epsilon =0.5$. Auto-PGD run used the same settings. For pixel attack, we use its variant combined with a threshold attack while employing covariance matrix adaptation evolution strategy (CMA-ES) instead of differential evolution (DE) algorithm as we found that DE is simply too slow and known to be worse at local solution finding. For ZOO we use learning rate $l =0.01$ and importance sampling.

\begin{table*}[htbp]
\centering
\begin{tabular}{|c|c|c|c|c|c|c|c|c|c|c|}
\hline
\multicolumn{11}{|c|}{\textbf{TCGA Data}} \\
\hline
\multicolumn{1}{|c|}{\textbf{Attack}} & \multicolumn{2}{c|}{\textbf{RF}} & \multicolumn{2}{c|}{\textbf{XGBoost}} & \multicolumn{2}{c|}{\textbf{ResNet}} & \multicolumn{2}{c|}{\textbf{CNN}} & \multicolumn{2}{c|}{\textbf{ViT}}\\
\hline
 & \textbf{Acc}$\downarrow$& \textbf{SSIM}$\uparrow$& \textbf{Acc}$\downarrow$& \textbf{SSIM}$\uparrow$& \textbf{Acc}$\downarrow$& \textbf{SSIM}$\uparrow$& \textbf{Acc}$\downarrow$& \textbf{SSIM}$\uparrow$& \textbf{Acc}$\downarrow$& \textbf{SSIM}$\uparrow$\\
\hline
FGSM          &  -  & - & - & - &  0.081 &  0.893&  0.001 & 0.851 &  0.186& 0.878 \\
Auto-PGD      & -  & - & - & - &  0.001&  0.986&  0.002&  0.985 &  0.121&  0.963 \\
ZOO*          &  0.994 &   0.999&  0.973&   0.999&  0.995&   0.999& 0.992 & 0.999&  0.995&   0.999\\
Pixel Attack* &   0.992&   0.999&  0.994&   0.999&  0.987&   0.999&  0.993& 0.999  &  0.992&  0.999 \\
\hline

\multicolumn{11}{|c|}{\textbf{COVID Data}} \\

\hline

\multicolumn{1}{|c|}{\textbf{}} &\multicolumn{2}{c|}{\textbf{RF}} & \multicolumn{2}{c|}{\textbf{XGBoost}} & \multicolumn{2}{c|}{\textbf{ResNet}} & \multicolumn{2}{c|}{\textbf{CNN}} & \multicolumn{2}{c|}{\textbf{ViT}} \\

\hline
 & \textbf{Acc}$\downarrow$& \textbf{SSIM}$\uparrow$& \textbf{Acc}$\downarrow$& \textbf{SSIM}$\uparrow$& \textbf{Acc}$\downarrow$& \textbf{SSIM}$\uparrow$& \textbf{Acc}$\downarrow$& \textbf{SSIM}$\uparrow$& \textbf{Acc}$\downarrow$& \textbf{SSIM}$\uparrow$\\
 \hline

FGSM            & - & - & - & - &  0.002&  0.333&  0.001&  0.428&  0.001& 0.397 \\
Auto-PGD        & -  & - & - & - &  0.001&  0.681&  0.003&  0.681&  0.323&   0.680 \\
ZOO*            &   0.974&  0.998&  0.992&  0.998&  0.916&  0.998&  0.782&  0.998&  0.889& 0.998\\
Pixel Attack*   &   0.979&  0.998& 0.993 &  0.998&  0.986&  0.998&  0.902&  0.998&  0.968&  0.998\\

\hline

\end{tabular}
\caption{Comparative analysis of different attacks across all models. Acc stands for accuracy after the attack. * we found that black-box methods while carefully perturbing the data to avoid detection, ultimately failed to significantly affect it and produce a good result}
\label{methodcomparison_extra}
\end{table*}

In table \ref{methodcomparison_extra} we note that while on a simpler TCGA dataset, we outperform black-box models only in terms of accuracy after the attack while maintaining competitive SSIM, thus indicating the undetectability of our attack, on COVID-19 dataset we outperform even white-box methods on undetectability. We postulate this being because COVID-19 is a complex dataset and while we rely on feature importance, other methods rely on directly or indirectly estimating a gradient of a model. 
Additionally, from our experiments, we observed that pixel attack in particular is extremely slow on the transformer on high iteration settings.
We note that overall attacks on images are very inefficient for trees and are difficult to adapt as you cannot batch the input as effectively as on deep learning models. Overall the results indicate better performance compared to popular black-box approaches, and competitive undetectability of our method compared with white-box attacks.

\subsection{Synthetic Data}
One limitation of all adversarial attacks can be the access to data, to address this we test our VAE model on the task of generating poisonous data.
To verify the ability of our VAE to create believable synthetic data, we calculate MSE, cosine similarity, and structural similarity index (SSIM), the latter being the commonly preferred way of verifying the reconstruction of images, as with this case we treat gene expression vectors as one channel images and flatten the image for other metrics. We also extract a new metric:  TTG (time to generation in seconds) - how fast our attack can generate n poisonous samples, here $n=1000$. The summary is shown in table \ref{vae-table}.
\begin{table}
   \centering
  \begin{tabular}{|l|l|l|l|l|}
    \hline
    \textbf{VAE Model}     &\textbf{TTG} $\downarrow$ & \textbf{MSE} $\downarrow$& \textbf{SSIM} $\uparrow$ & \textbf{Cos} $\uparrow$  \\
    \hline
    VAE-TCGA     & 44.74s  & 0.205  & 0.545  &  0.779\\
    VAE-COVID     & 43.04s &   0.047 & 0.406  & 0.866  \\
    
    \hline
  \end{tabular}
   \caption{VAE performance metrics}
  \label{vae-table}

\end{table}
This suggests that attackers can via a relatively straightforward method rapidly generate toxic synthetic data, this poses a particularly serious danger for future open-source data pipelines, as most genomics datasets are released with open licenses and available for download without any verification. They in turn can be modified and reuploaded, models trained on such datasets are therefore more likely to be vulnerable to adversarial attacks. The adjacent benefit of this is the creation of synthetic adversarial data with correct labeling for safety training to improve the robustness of the underlying model.

\subsection{Detection and Countermeasures}
\begin{figure}[htbp]
\centering
    \includegraphics[scale=0.13]{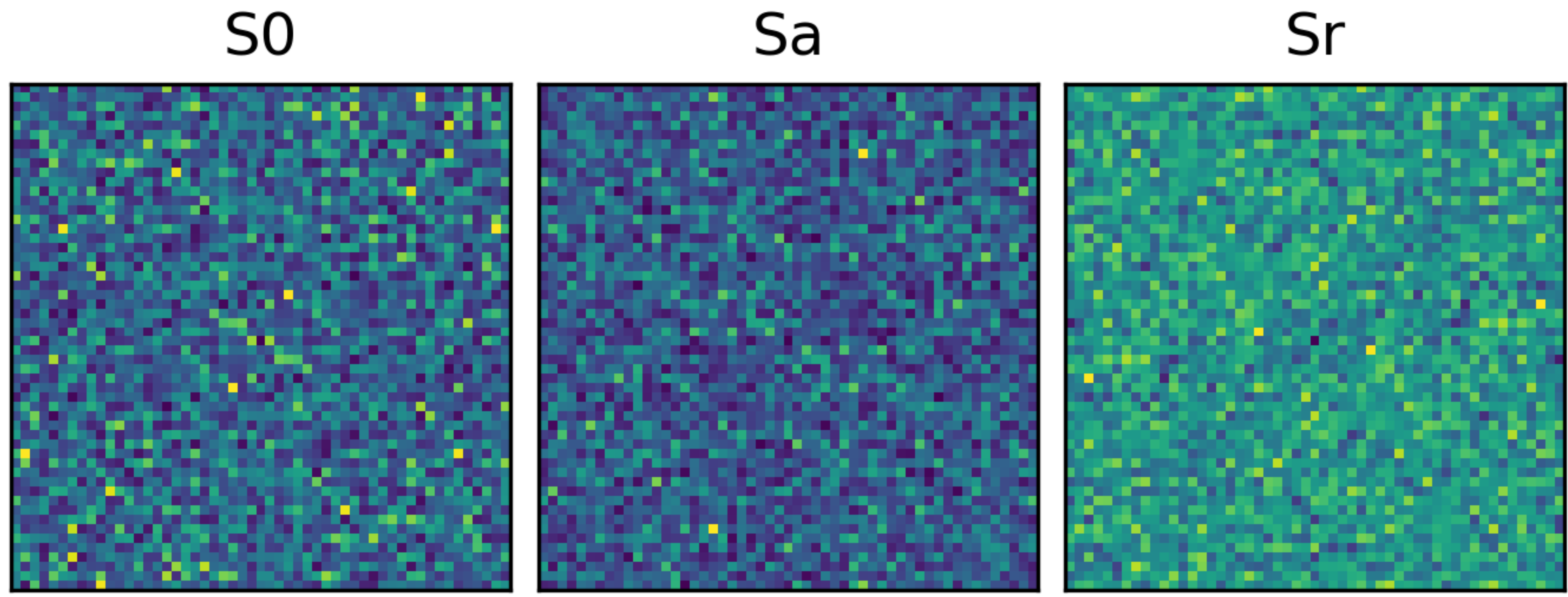}
    \caption{Example of spectral analysis of a ResNet model performed on TCGA dataset: S0 - original dataset, Sa - transformed dataset (main method), Sr transformed dataset (brute force method) }
    \label{fig:fftexample}
\end{figure}
To see the difference over the entire dataset, we treat gene expression values in a sample as a signal, we then combine the samples into a single signal and compute a fast Fourier Transform (FFT) to identify the power spectrum difference between the attacked and original datasets for each model at maximum severity setting of the attack. This is a widely used defense method for analyzing and detecting attacks in computer vision \cite{bafna2018thwarting,harder2021spectraldefense,yahya2020probabilistic,xu2020lance}, therefore utilize it as our detection method to measure how well our attack can stay hidden. After extracting 2D spectrum plots similar to figure \ref{fig:fftexample}, we compute SSIM between the original dataset spectrum and the attacked versions, we do so also on the subtests of vulnerable data. The results of this experiment presented in table \ref{methodcomparison} unequivocally demonstrate the benefit of our main approach over brute force, with attacked datasets received from the main attack closely mimicking the original data, while randomized perturbations are easily picked up by the detection method.
We postulate that the following steps can improve the robustness of future and existing pipelines:
\begin{itemize}
    \item Studying the complexity of the dataset in the context of model training can help reduce the susceptibility of the model to failure of detection of FP/FN samples, we want to encourage further studies into the effects of data topology on the effectiveness of adversarial attacks.
    \item Detecting and keeping track of vulnerable subsets by pre-attacking the model, this step is crucial as currently there are no widely developed and accepted practices of applying adversarial analysis and safety training to models trained on genomics data. Having data on vulnerability analysis can illuminate model shortcomings that can be addressed via either adversarial training or an additional trained detector.
    \item Development of more sophisticated defense methods. While the computer vision domain boasts a plethora of defense methods that can be taken as inspiration, we suggest that their variations when applied to the field of AI in genomics have to be carefully adjusted and validated as they might be ineffective due to the underlying difference in feature topology.
\end{itemize}

\section{Conclusions}

We introduced a new adversarial robustness framework that successfully attacked popular state-of-the-art machine learning models that were trained on genomic data of different levels of complexity. Our method is a simple black-box approach that does not rely on model gradients and can instead utilize fast iterative feature importance extraction with selective perturbations, demonstrating the impact of small data perturbations, which proves the vulnerability of current popular approaches.  We further extend our attack by developing a generative VAE model capable of rapidly creating believable poisonous data.
Further, we utilize spectral analysis as a popular AML defense method in computer vision to showcase the ability of our attack to mask adversarial samples as natural model error, while additionally discussing the potential countermeasures.
In the future, we hope to increase the effectiveness of attacks on ML models using a smaller combination of modified features akin to one-pixel attacks.  Moreover, we want to further explore generative attack methods and more sophisticated defense methods aimed at strengthening AI methods.
We hope that this work encourages researchers to carefully consider the robustness of their core AI pipelines in areas of genomics and on the clinical level. In addition, we want to further encourage active research into adversarial attacks on AI models in genomics.

\section{Code Availability}
The code demonstrating the implementation of the methods used in this paper, as well as dataset preprocessing for this paper will be made available in a GitHub repository here: 
\url{https://github.com/HeorhiiS/fimba-attack}

\section{Data Availability}
We kindly ask our readers to refer to our section \ref{sec:data} and appendix \hyperref[appendix:a]{A} for details on data acquisition and processing.
For COVID-19 dataset the files could be found here: \url{https://www.ncbi.nlm.nih.gov/geo/query/acc.cgi?acc=GSE171524}.

For TCGA dataset you can follow the download instructions here: \url{https://xenabrowser.net/datapages/?cohort=TCGA%20TARGET%20GTEx}.

\bibliography{paper}
\bibliographystyle{icml2023}

%%%%%%%%%%%%%%%%%%%%%%%%%%%%%%%%%%%%%%%%%%%%%%%%%%%%%%%%%%%%%%%%%%%%%%%%%%%%%%%
%%%%%%%%%%%%%%%%%%%%%%%%%%%%%%%%%%%%%%%%%%%%%%%%%%%%%%%%%%%%%%%%%%%%%%%%%%%%%%%
% APPENDIX
%%%%%%%%%%%%%%%%%%%%%%%%%%%%%%%%%%%%%%%%%%%%%%%%%%%%%%%%%%%%%%%%%%%%%%%%%%%%%%%
%%%%%%%%%%%%%%%%%%%%%%%%%%%%%%%%%%%%%%%%%%%%%%%%%%%%%%%%%%%%%%%%%%%%%%%%%%%%%%%
\newpage
\appendix
\onecolumn

\section*{Appendix A: Data Extraction and Preprocessing}\label{appendix:a}
\begin{figure}[!htb]
 
    \centering
    \includegraphics[scale=0.5]{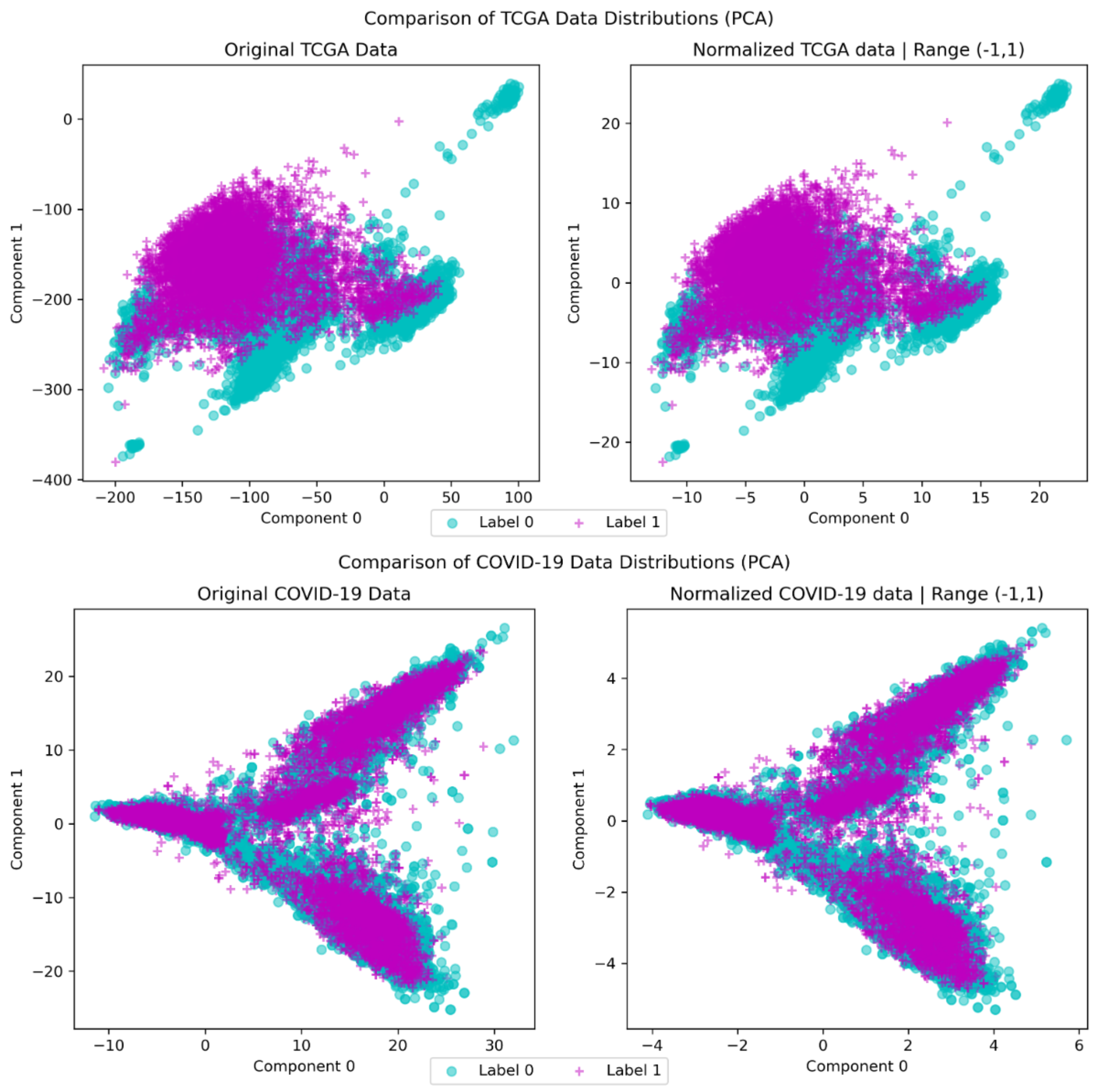}
    % \fbox{\rule[-.5cm]{0cm}{4cm} \rule[-.5cm]{4cm}{0cm}}
    \caption{Distribution of datasets before and after normalization using PCA (components=100, top 2 plotted)}
    \label{fig:pcadist}
\end{figure}

Our primary steps for data processing included normalizing if required, relabeling, selection, and cleaning.
All steps were performed on an HPC platform utilizing 10 AMD EPYC 7742 64-Core CPU processors when parallel computing was available. Cleaning was performed using Pandas and Numpy Python libraries. For the TCGA-TARGET-GTEx cohort, an additional step of gene filtering was required and was completed utilizing the popular bioinformatics \textit{Scanpy} library using a Cell Ranger configuration that is developed by 10x Genomics, the same platform that was used to process our COVID-19 dataset. We apply PCA to both datasets before and after normalization to verify their distributional integrity as well as to get a better understanding of the structure of the data. Figure \ref{fig:pcadist} demonstrates that the structure of data is retained after normalization, as well as that, upon observation, COVID-19 data has a higher complexity, which can explain lowe

\subsection*{Cancer Data From TCGA Dataset}
The raw TCGA-TARGET-GTEx data was downloaded from Xena Browser - a data hosting and analysis repository provided by the University of California Santa Cruz. We chose RSEM FPKM RNA-seq gene expression data as our dataset of choice. This setting meant that we are extracting processed data collected using the RSEM software package for estimating gene and isoform expression levels from single-end or paired-end RNA-Seq data and that it was normalized using the FPKM (fragments per kilobase of transcript per million reads mapped) method. As gene counts in raw count sparse matrices are often represented as large integers, an additional normalization step is standard. Upon initial data exploration, the data contained 19131 samples with 60498 features being specific gene types demonstrating their counts per cell. To relabel and reorient the data we additionally collected two more databases from Xena Browser. The first is a "probemap", a collection of mappings between Ensembl gene ID types e.g.(ENSG00000227232.5) to a common symbol e.g.(WASH7P). The next set was a phenotype map, that contained sample IDs, a detailed category or cancer type, a primary site of cancer, sample type (tumor or normal), the gender of a patient from whom the sample was taken, and the study (TCGA,TARGET,GTEx). 

As not all of the 60498 genes may carry meaningful differential expression data we applied widely used filtering techniques to select features capable of quality representation. First, we utilized the \textit{scprep} package to remove rare, unrepresentative genes. This step filtered all genes with negligible counts in all but a few cells. This allowed us to reduce the number of features from 60498 to 41057. Next, to further increase the quality of our data we used a common approach of selecting highly variable genes using the \textit{Scanpy} package. By loading our data into an AnnData object we applied a function to perform the selection, by setting the minimum mean at 0.01, and maximum mean at 8, using cell ranger as a selector, and binning the results in 20 bins, we managed to select 2881 genes out of 41057, further increasing the quality of our data. Then we mapped sample IDs to labels and converted the sample type into a binary label of 0 or 1, where class types in a set: {'Primary Tumor',  'Recurrent Tumor',
       'Metastatic', 'Additional Metastatic',
       'Primary Blood Derived Cancer - Peripheral Blood',
       'Recurrent Solid Tumor', 'Primary Solid Tumor',
       'Recurrent Blood Derived Cancer - Bone Marrow',
       'Primary Blood Derived Cancer - Bone Marrow',
       'Recurrent Blood Derived Cancer - Peripheral Blood'} were given a label of 1 and classes in {'Solid Tissue Normal' 'Normal Tissue'} were given a label 0. We then matched the core dataset on sample ID, then an additional final filtering step was performed to make sure that the dataset doesn't contain NaN values or artifacts. 

\subsection*{COVID-19 Data}

COVID-19 as a viral type was selected due to a large amount of available data, increased industry adoption of tools to predict and detect the infection, and recent urgency and importance of COVID-19-related analysis. A search and filtering procedure on NCBI GeoBank yielded a dataset with an accession ID of GSE171524. The partially processed version of this dataset was downloaded to keep the original method of filtering intact. Since the dataset and the original gene matrices were filtered to contain only the genes with the highest quality and representational ability, no gene selection step was required. 

Additionally, a metadata map containing sample IDs, labels, and cell types was extracted from NCBI to allow for accurate label mapping. The dataset and map were matched on the sample ID, and 19 cell types were given a unique categorical label. Group columns were mapped to binary labels 0 and 1, corresponding to the negative and positive cases of COVID-19. After an additional filtering step to ensure no NaNs or artifacts from processing steps was done the resulting dataset contained 116314 samples and 2000 features per sample. Both datasets were also normalized in the range between $R=(-1,1)$ as is the common practice and to prepare it for deep learning. Due to the size of the dataset, we uniformly sampled 50000 samples using random seed $= 42$.

%%%%%%%%%%%%%%%%%%%%%%%%%%%%%%%%%%%%%%%%%%%%%%%%%%%%%%%%%%%%%%%%%%%%%%%%%%%%%%%
%%%%%%%%%%%%%%%%%%%%%%%%%%%%%%%%%%%%%%%%%%%%%%%%%%%%%%%%%%%%%%%%%%%%%%%%%%%%%%%
\clearpage
\section*{Appendix B: Additional VAE Information}
\subsection*{VAE Architecture and Sample Reconstruction }
Here we present a diagram of VAE architecture to provide visually expanded information on the referenced model in the paper. Note that we arrived at this architecture after an extensive hyperparameter search for the best configuration. For the activation function, we use hyperbolic tangent. For the latent dimension, we employ n=500, and for the input dimension we utilize either 2880 or 2000 depending on the dataset type (TCGA or COVID-19) respectively. We also include an example of reconstructed output after interpolation taken on a range of $(0,100)$, at $n-1$, hence $99$. We note that the reconstructed sample while possessing some differences, generally replicates the original in terms of the spectrum presented in the example.

\begin{figure}[htb]
 
    \centering
    \includegraphics[scale=0.48]{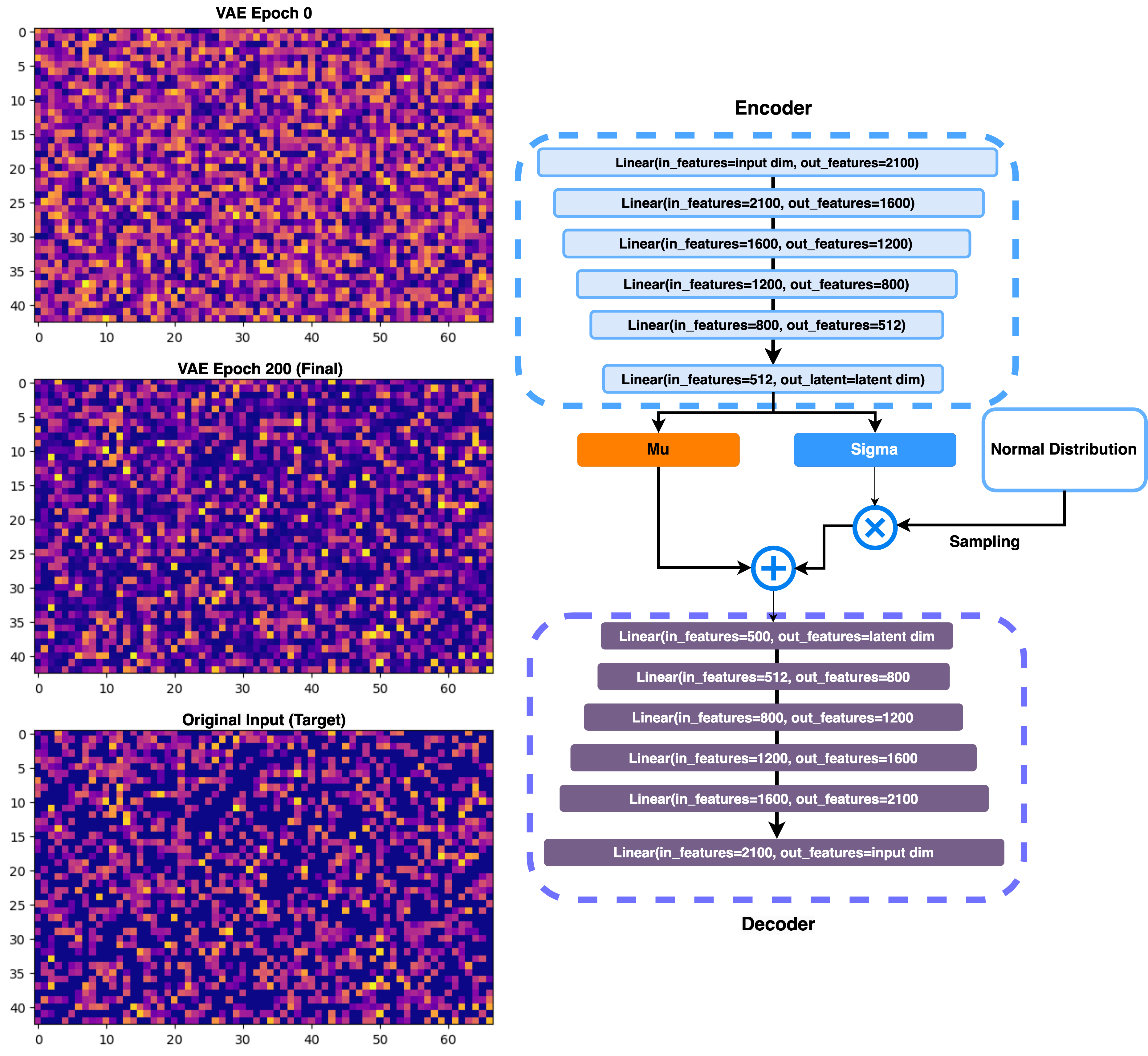}
    % \fbox{\rule[-.5cm]{0cm}{4cm} \rule[-.5cm]{4cm}{0cm}}
    \caption{VAE generated samples and architecture}
    \label{fig:vae_info}
\end{figure}

\section*{Appendix C: Additional Results}
\subsection*{TP/TN Endpoint variant of the Main Attack}
As mentioned in the paper, we experimented with different settings for the attack, instead of choosing a sample that reduces model confidence like a false positive or false negative as an interpolation endpoint we tried a similar approach but chose true positives and true negatives instead. Our results below indicate that even in case of lack of knowledge of which sample is a false positive or false negative, an attacker can transform the samples to mimic natural model error with samples of opposite class to the original sample to be attacked. The selection of a pair can be done by querying the model until the samples with two opposite labels are selected. This further validates our black box approach. On a further note, this attack was particularly effective on the XGBoost model as shown below.
\begin{figure}[htb]
 
    \centering
    \includegraphics[scale=1]{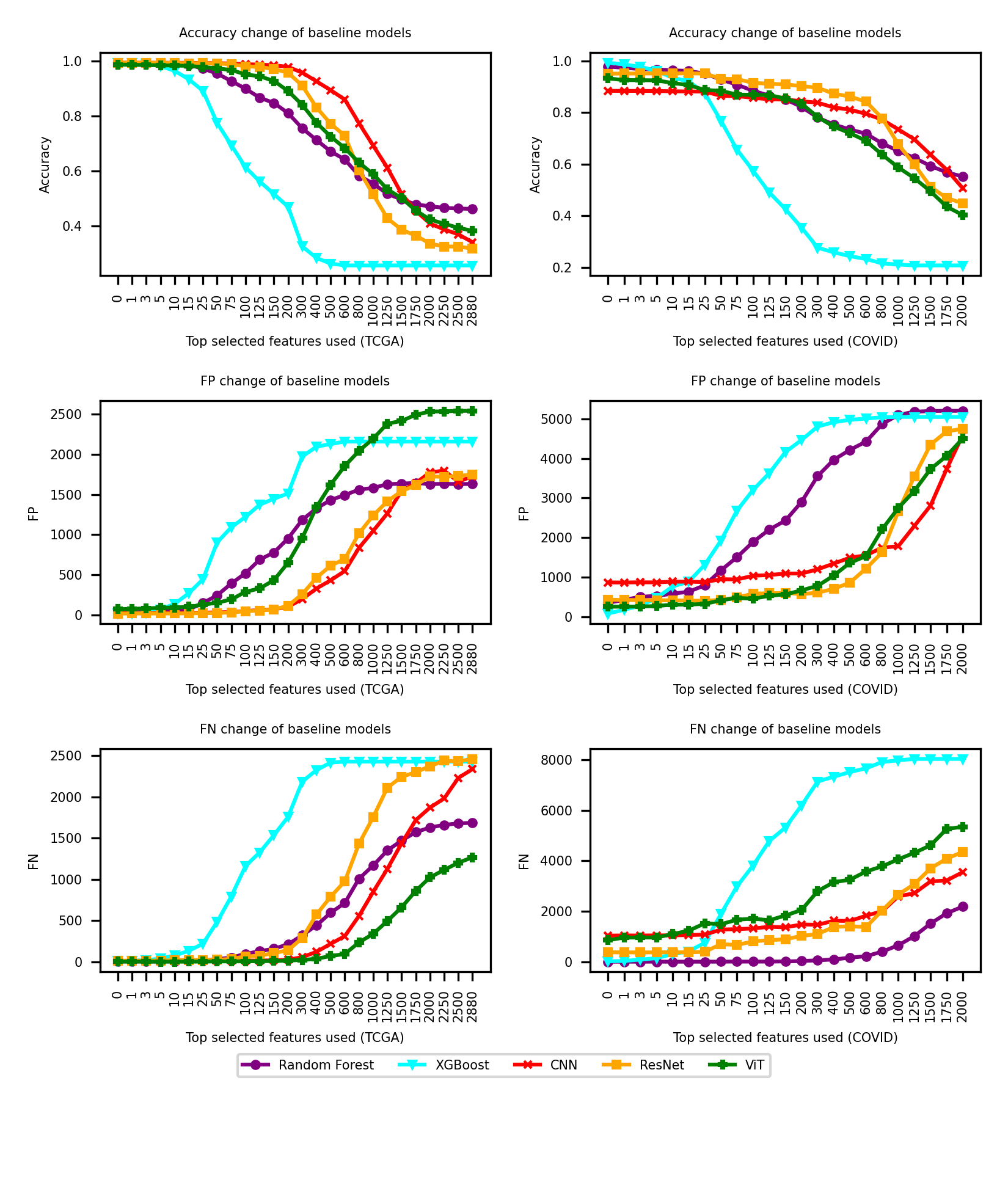}
    % \fbox{\rule[-.5cm]{0cm}{4cm} \rule[-.5cm]{4cm}{0cm}}
    \caption{Results of our main attack using TN/TP as interpolation endpoint (transformation target)}
    \label{fig:tntp_res}
\end{figure}
\end{document}